\def\year{2022}\relax
\documentclass[letterpaper]{article} 
\usepackage{aaai22}  
\usepackage{times}  
\usepackage{helvet}  
\usepackage{courier}  
\usepackage[hyphens]{url}  
\usepackage{graphicx} 
\usepackage{natbib}  
\usepackage{caption} 
\DeclareCaptionStyle{ruled}{labelfont=normalfont,labelsep=colon,strut=off} 
\usepackage{graphicx}
\usepackage{amsmath}
\usepackage{amssymb}
\usepackage{multirow}
\usepackage{makecell}
\usepackage{bbding}
\usepackage{bm}

\usepackage{algorithm}
\usepackage{algorithmic}

\usepackage{newfloat}
\usepackage{listings}
\floatstyle{ruled}
\newfloat{listing}{tb}{lst}{}
\floatname{listing}{Listing}

\usepackage{subfig}
\usepackage{array}

\newcolumntype{x}[1]{>{\centering\arraybackslash}p{#1pt}}
\newcommand{\tablestyle}[2]{\setlength{\tabcolsep}{#1}\renewcommand{\arraystretch}{#2}\centering\footnotesize}

\usepackage[switch]{lineno}
\usepackage{color}

\newcommand{\ie}{\textit{i}.\textit{e}.}
\newcommand{\eg}{\textit{e}.\textit{g}.}
\newcommand{\cf}{\textit{cf.}}

\newcommand{\vs}{\textit{vs}.}

\title{Rethinking the Two-Stage Framework for Grounded Situation Recognition}

\author {
    Meng Wei,\textsuperscript{\rm 1}\thanks{Work started when M. Wei and L. Chen at Tencent AI Lab.}
    Long Chen,\textsuperscript{\rm 2}\footnotemark[1]\thanks{Corresponding authors.}
    Wei Ji,\textsuperscript{\rm 1}\footnotemark[2]
    Xiaoyu Yue,\textsuperscript{\rm 3}
    Tat-Seng Chua\textsuperscript{\rm 1}\\
}
\affiliations {
    \textsuperscript{\rm 1} Sea-NExT Joint Lab, National University of Singapore \\
    \textsuperscript{\rm 2} Columbia University \\
    \textsuperscript{\rm 3} Centre for Perceptual and Interactive Intelligence \\
    \{kellymeng0427, zjuchenlong, yuexiaoyu002\}@gmail.com, jiwei@nus.edu.sg, chuats@comp.nus.edu.sg
}

\begin{document}


\maketitle

\begin{abstract}
Grounded Situation Recognition (GSR), \ie, recognizing the salient activity (or verb) category in an image (\eg, \texttt{buying}) and detecting all corresponding semantic roles (\eg, \texttt{agent} and \texttt{goods}), is an essential step towards ``human-like" event understanding. Since each verb is associated with a specific set of semantic roles, all existing GSR methods resort to a two-stage framework: \emph{predicting the verb in the first stage and detecting the semantic roles in the second stage.} However, there are obvious drawbacks in both stages: 1) The widely-used cross-entropy (XE) loss for object recognition is insufficient in verb classification due to the large intra-class variation and high inter-class similarity among daily activities. 2) All semantic roles are detected in an autoregressive manner, which fails to model the complex semantic relations between different roles. To this end, we propose a novel \textbf{SituFormer} for GSR which consists of a Coarse-to-Fine Verb Model (CFVM) and a Transformer-based Noun Model (TNM). CFVM is a two-step verb prediction model: a coarse-grained model trained with XE loss first proposes a set of verb candidates, and then a fine-grained model trained with triplet loss re-ranks these candidates with enhanced verb features (not only separable but also discriminative).
TNM is a transformer-based semantic role detection model, which detects all roles parallelly.
Owing to the global relation modeling ability and flexibility of the transformer decoder, TNM can fully explore the statistical dependency of the roles.
Extensive validations on the challenging SWiG benchmark show that SituFormer achieves a new state-of-the-art performance with significant gains under various metrics.
Code is available at \url{https://github.com/kellyiss/SituFormer}.

\end{abstract}

\begin{figure}[t]
	\centering
	\includegraphics[width=0.9\linewidth]{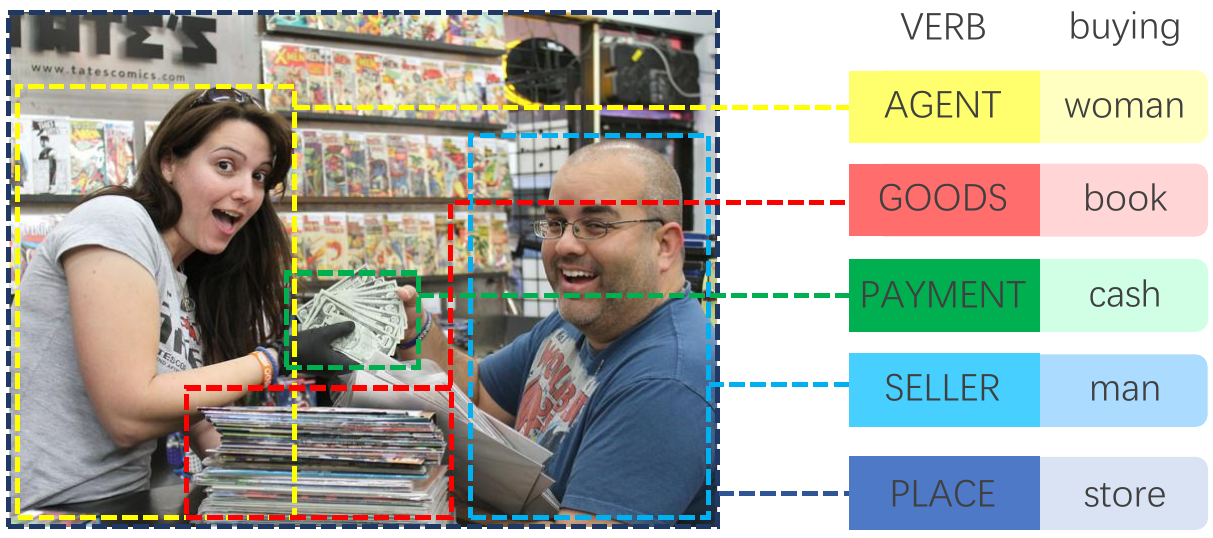}
	
	(a)
	
	\includegraphics[width=0.9\linewidth]{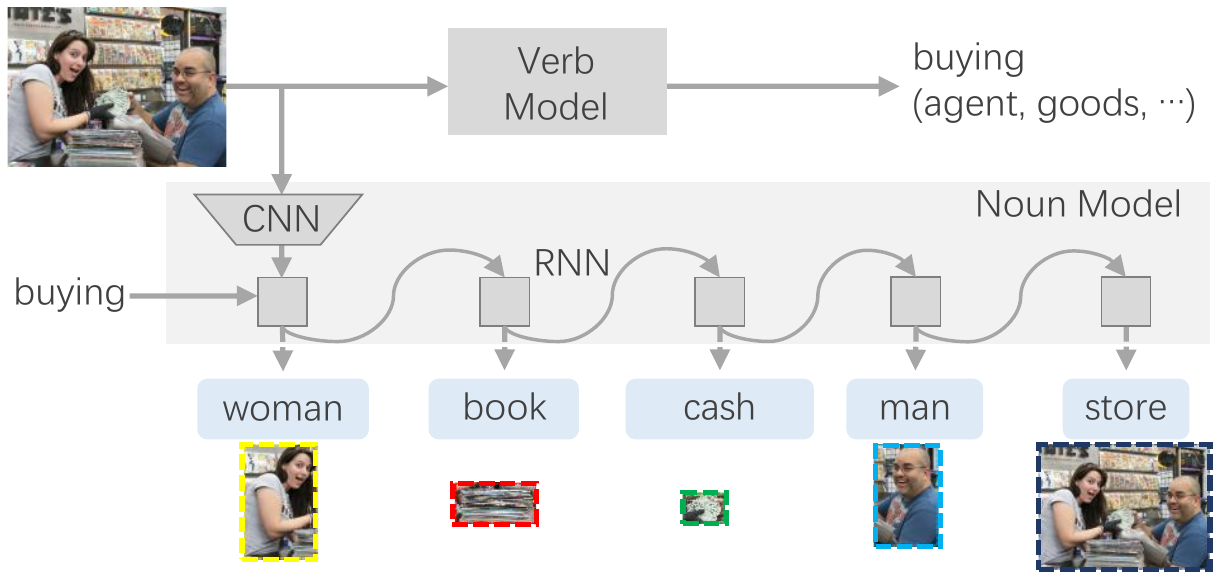}
	
	(b)
	\caption{(a) An example of GSR. Given this image, a GSR model needs to not only predict the verb category \texttt{buying}, but also detect (\ie, classify and ground) all corresponding semantic roles for \texttt{buying} event, such as \texttt{agent}, \texttt{goods}, and \texttt{payment}. (b) An overview of the existing two-stage GSR framework, which consists of a verb model and an RNN-based noun model to detect all roles autoregressively.
	}
	\label{fig:motivation}
\end{figure}

\section{Introduction}




Understanding activities in images is one of the core tasks for computer vision. 
With the maturity of action recognition~\cite{carreira2017quo, wang2016temporal} and object detection~\cite{ren2015faster}, today's computers can recognize action or object categories well. 
However, ``human-like" activity understanding goes beyond action-centric or object-centric recognition. 
A more crucial step is to identify how objects participate in activities, such as ``\textsc{someone do something with some-tool at someplace}". 
Hence, the task \textbf{Situation Recognition (SR)}~\cite{yatskar2016situation} is proposed for comprehensive event extraction.  
As the example in Figure~\ref{fig:motivation} (a), SR not only recognize the salient activity(or verb) in the image (\eg, \texttt{buying}), but also recognize all semantic roles (\eg, \texttt{agent} is \texttt{woman}, \texttt{place} is \texttt{store}). To further ground the semantic roles in the image, a more challenging task \textbf{Grounded Situation Recognition (GSR)}~\cite{pratt2020grounded} was proposed (\cf~bounding boxes in Figure~\ref{fig:motivation} (a)). By describing activities with verb and grounded semantic roles, GSR provides a visually-grounded structure representation (named verb frame) for the activity, which benefits many downstream scene understanding tasks, such as image-text retrieval~\cite{gordo2016deep, noh2017large}, image captioning~\cite{mallya2017recurrent,chen2021human,chen2017sca}, visual grounding~\cite{chen2021ref}, and VQA~\cite{cadene2019murel,Chen_2020_CVPR,chen2021counterfactual,xiao2022video}.

Since each verb is inherently associated with a specific set of semantic roles (\eg, semantic role set $\langle$\texttt{agent}, \texttt{goods}, \texttt{payment}, \texttt{seller}, \texttt{place}$\rangle$ for verb \texttt{buying}), almost all existing SR methods resort to a \emph{two-stage} framework: 1) predicting the verb (or action categories) for the whole image in the first stage; and 2) predicting nouns (or object categories) for all semantic roles in the second stage. Inspired by the success of SR methods, state-of-the-art GSR methods also follow the same two-stage framework by replacing the second-stage role classification model with a semantic role detection model. To the best of our knowledge, there are two existing SOTA GSR models: ISL and JSL~\cite{pratt2020grounded}. Specifically, as summarized in Figure~\ref{fig:motivation} (b), they are all two-stage models. For verb prediction, they train a verb model with N-way cross-entropy (XE) loss. For semantic role detection, they utilize an RNN-based noun model to predict and ground the noun for each semantic role autoregressively, \ie, they feed the predicted noun embedding of the last semantic role back into the RNN to guide the next prediction.

\begin{figure}[t]
    \centering
    \includegraphics[width=0.98\linewidth]{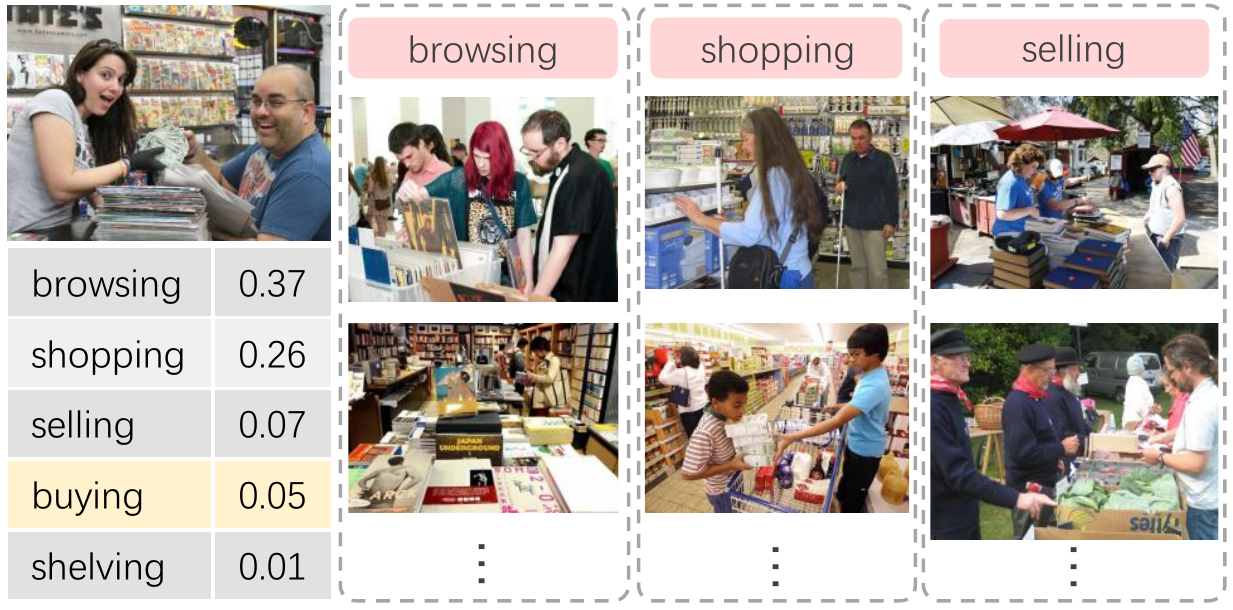}
    \caption{\textbf{Left}: A failure example of the verb model trained with XE loss and its predicted verb distributions. Its ground-truth verb is \texttt{buying}. \textbf{Right}: Some randomly selected images from the training set of the same category of hard negative verbs (\ie, \texttt{browsing}, \texttt{shopping}, and \texttt{selling}).}
    \label{fig:rerank_intro}
\end{figure}

Although existing two-stage GSR methods have achieved satisfactory performance, we argue that there are still some unreasonable designs in both two stages:

\noindent\textbf{Verb Model} (the first-stage): Since each verb can have different combinations of nouns w.r.t the semantic role set, the activity patterns are much more complex than objects (\ie, larger intra-class variation and higher inter-class similarity). Thus, even using a deep ConvNet  (\eg, ResNet-50~\cite{he2016deep}) trained with XE loss can still fail to discriminate ambiguous verbs which place emphasis on different semantic roles. For example, in Figure~\ref{fig:rerank_intro}, due to the frequent occurrences of ``people browsing books at bookstore" and similar scene appearance (\cf~images of \texttt{browsing}), the test image is tended to be wrongly predicted as \texttt{browsing}. Instead, if the verb model can focus more on some discriminative roles (\eg, the \texttt{payment} is happening with \texttt{cash} in \texttt{hands}), it would be easier to distinguish the \texttt{buying} from these plausible verb choices.

\noindent\textbf{Noun Model} (the second-stage): 1) RNN-based models simply formulate each situation as a sequence of semantic roles, \ie, this link structure fails to model the complex relations between different semantic roles. 2) This autoregressive sequential prediction manner is prone to result in error accumulation. 3) They only utilize noun category embeddings to guide the training, which is easier to suffer from severe semantic sparsity issue~\cite{yatskar2017commonly}, especially when the number of noun categories is extremely large (\eg, $\approx$ $10,000$ categories in SWiG benchmark).

In this paper, to address the above-mentioned issues, we propose a novel two-stage model (\ie, a verb model and a noun model): Situation Transformer (dubbed \textbf{SituFormer}).

For the verb model, since the verb feature learned with XE loss is not discriminative enough, we enhance it using triplet loss with a carefully designed hard triplet mining scheme. Similar practice are common in face recognition~\cite{schroff2015facenet,wen2016discriminative}. Specifically, it is a coarse-to-fine two-step model. In the coarse-grained step, we first predict the \emph{top-N} verbs with a coarse-grained verb model trained by XE loss. Then, in the fine-grained step, we mine hard triplets from images of the \emph{top-N} verbs considering the semantic role feature similarity. After further training a lightweight fine-grained model with triplet loss to obtain effectively enhanced verb features of all training samples (the gallery), the fine-grained model can re-rank \emph{top-N} verbs by considering the feature similarity with the support image samples from the gallery.



For the noun model, it is a transformer-based encoder-decoder model. The input queries for the decoder are a set of learnable embeddings for a verb and its corresponding semantic roles. The outputs of the decoder for each query are the predicted object category and grounding location. The built-in self-attention mechanism in the decoder implicitly formulates each verb frame as a fully-connected graph structure (\vs~the sequence structure in existing GSR models). Meanwhile, our parallel decoding paradigm can avoid error accumulation. Moreover, sharing semantic role query embeddings across different verb frames introduces useful inductive bias, which alleviates the semantic sparsity issue.

We evaluate our proposed SituFormer on the challenging GSR benchmark: SWiG~\cite{pratt2020grounded}. Extensive experiments have demonstrated the effectiveness of each component. Without bells and whistles, SituFormer outperforms all state-of-the-art GSR models on all evaluation metrics.

\section{Related Work}
\noindent\textbf{Situation Recognition (SR).}
SR is first proposed by~\cite{gupta2015visual,yatskar2016situation}, which generalizes action classification~\cite{carreira2017quo,9151084}, HOI~\cite{liao2020ppdm,zou2021end,wei2020hose} and SGG~\cite{chen2019counterfactual,cong2021spatial}, and aims to provide a structured representation for an activity (event) with a verb frame. Typically, the verb frame consists of a verb with a specific semantic role set drawn from FrameNet~\cite{10.3115/980845.980860}. The early CRF-based SR methods~\cite{yatskar2016situation, yatskar2017commonly} jointly predict verb and nouns by structured learning. 
However, sharing the visual representation for the two tasks has been proved inferior to training separate models in a two-stage way~\cite{mallya2017recurrent}.
Hence, recent RNN-based methods~\cite{mallya2017recurrent}, GNN-based methods~\cite{li2017situation,suhail2019mixture} and attention-based methods~\cite{9156513} always predict the verb in the first stage and recognize the semantic roles in the second stage.

\noindent\textbf{Ground Situation Recognition (GSR).} GSR~\cite{yang2016grounded,silberer2018grounding,pratt2020grounded} extends the SR task and aims to further ground the semantic roles which is critical to visual reasoning. The two existing GSR methods (JSL and ISL)~\cite{pratt2020grounded} follow the RNN-based two-stage SR pipeline to first predict the verb and then sequentially detect the semantic roles.  
Our method differs in two aspects: 1) the verb prediction is in a coarse-to-fine manner. 2) the semantic roles are detected in parallel rather than in autoregressive sequence. Another concurrent work~\cite{cho2021grounded} also resorts to Transformer structure for GSR.

\section{Approach}
\subsection{Overview}
Given an image $I$, GSR aims to detect a structured visually-grounded verb frame $\mathcal{G} = \left \{v, \mathcal{R}, \mathcal{N}, \mathcal{B}\right \}$, where $v \in \mathcal{V}$ is the category of the salient activity (or verb) in image $I$, and $\mathcal{R} = \{r_1, ..., r_m \}$ is the set of manually predefined semantic roles\footnote{These predefined semantic roles can be easily retrieved from the verb lexicon such as PropBank or FrameNet.} for verb $v$. $\mathcal{N} = \{n_1, ..., n_m \}$ and $\mathcal{B} = \{b_1, ..., b_m\}$ are the set of object (or noun) categories and bounding boxes for all semantic roles, \ie, $n_i \in \mathcal{O}$ is the object category of semantic role $r_i$, and $b_i \in \mathbb{R}^4$ is the bounding box location of semantic role $r_i$. $\mathcal{V}$ and $\mathcal{O}$ denote the predefined ontology of all possible verb and noun categories, respectively.

Currently, almost all existing GSR (or SR) models decompose this task into two steps: verb classification and noun detection (or classification). Thus, for GSR:
\begin{equation}
p(\mathcal{G} | I) = \underbrace{p(v, \mathcal{R} | I)}_{\text{Verb classification}} \underbrace{p(\mathcal{N}, \mathcal{B} | v, \mathcal{R}, I)}_{\text{Noun detection}}.
\end{equation}

\begin{figure}[t]
    \centering
    \includegraphics[width=\linewidth]{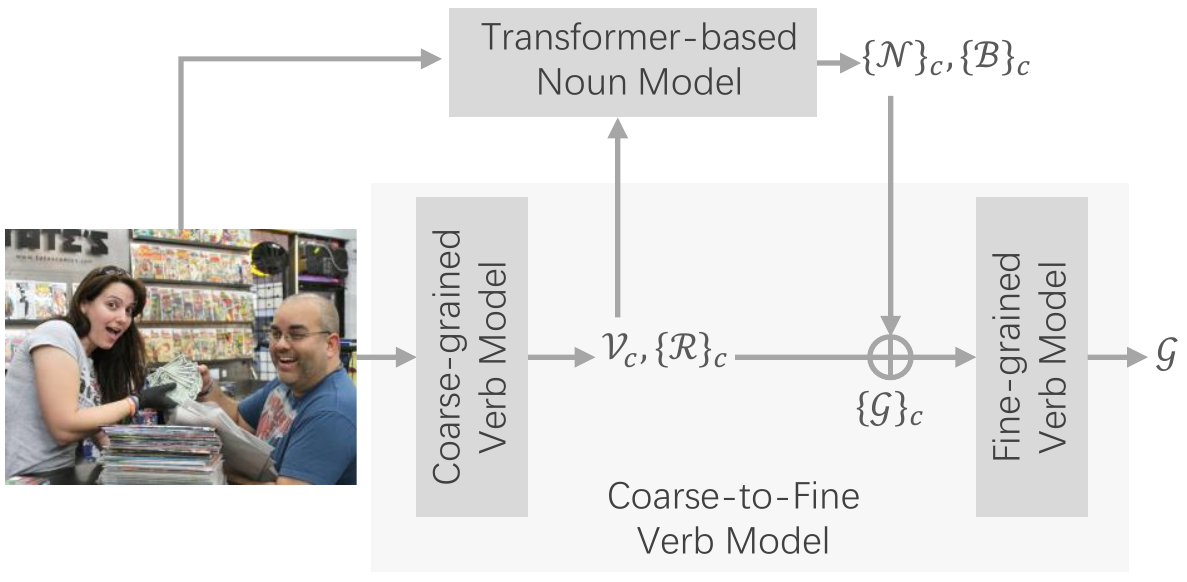}
    \caption{The overview pipeline of our SituFormer.}
    \label{fig:overview}
\end{figure}

In this paper, we propose a novel Situation Transformer (dubbed as SituFormer). It follows the same spirit and consists of two components: a Transformer-based Noun Model (TNM) and a Coarse-to-Fine Verb Model (CFVM). As illustrated in Figure~\ref{fig:overview}, given an image $I$, we first use a coarse-grained verb model (Verb-c) to propose a set of verb candidates (and their corresponding semantic roles), denoted as $\mathcal{V}_c$ and $\{\mathcal{R}\}_c$. Then, for all verb candidates, the TNM will output their respective noun categories $\{\mathcal{N}\}_c$ and bounding boxes $\{\mathcal{B}\}_c$. Lastly, a lightweight fine-grained verb model (Verb-f) selects the final verb frame prediction. Thus, we reformulate GSR as:
\begin{equation}
    \begin{aligned} \label{equ:2}
    & p(\mathcal{G} | I) = p (\{\mathcal{G}\}_c | I) p(\mathcal{G} |\{\mathcal{G}\}_c,I) \nonumber \\
    & =  p(\mathcal{V}_c, \{\mathcal{R}\}_c, \{\mathcal{N} \}_c, \{\mathcal{B}\}_c| I) p(\mathcal{G} |\{\mathcal{G}\}_c,I) \\
    & = 
    \underbrace{p(\mathcal{V}_c, \{\mathcal{R}\}_c| I)}_{\text{Verb-c}}
    \underbrace{p(\{\mathcal{N} \}_c, \{\mathcal{B}\}_c|\mathcal{V}_c, \{\mathcal{R}\}_c, I)}_{\text{TNM}}
    \underbrace{p(\mathcal{G} |\{\mathcal{G}\}_c,I) \nonumber}_{\text{Verb-f}},
\end{aligned}
\end{equation}
where $\{\mathcal{G}\}_c$ denotes the set of all verb frame candidates.

In this section, we first introduce each component of SituFormer, including TNM and CFVM (Verb-c and Verb-f). Then, we demonstrate the details of all training objectives.






\begin{figure*}
    \begin{minipage}{\textwidth}
        \begin{minipage}[t]{0.36\textwidth}
            \centering
            \includegraphics[width=0.9\linewidth]{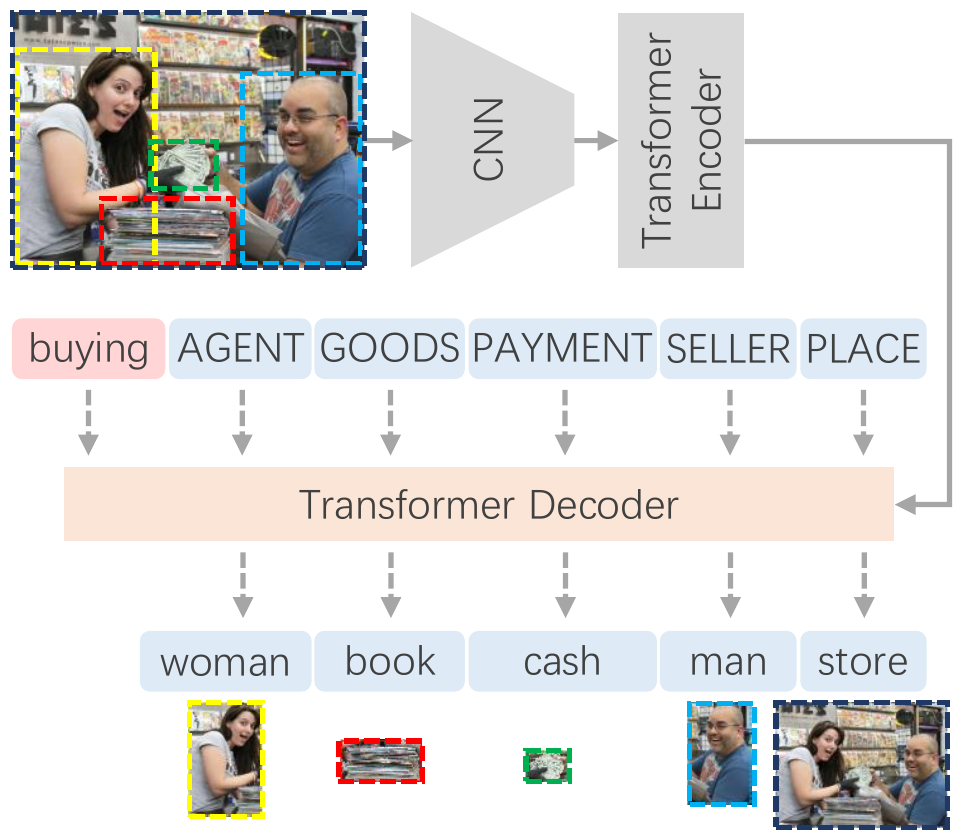}
            \caption{The architecture of TNM.}
            \label{fig:noun_model}
        \end{minipage}
        \begin{minipage}[t]{0.64\textwidth}
             \centering
            \includegraphics[width=0.9\linewidth]{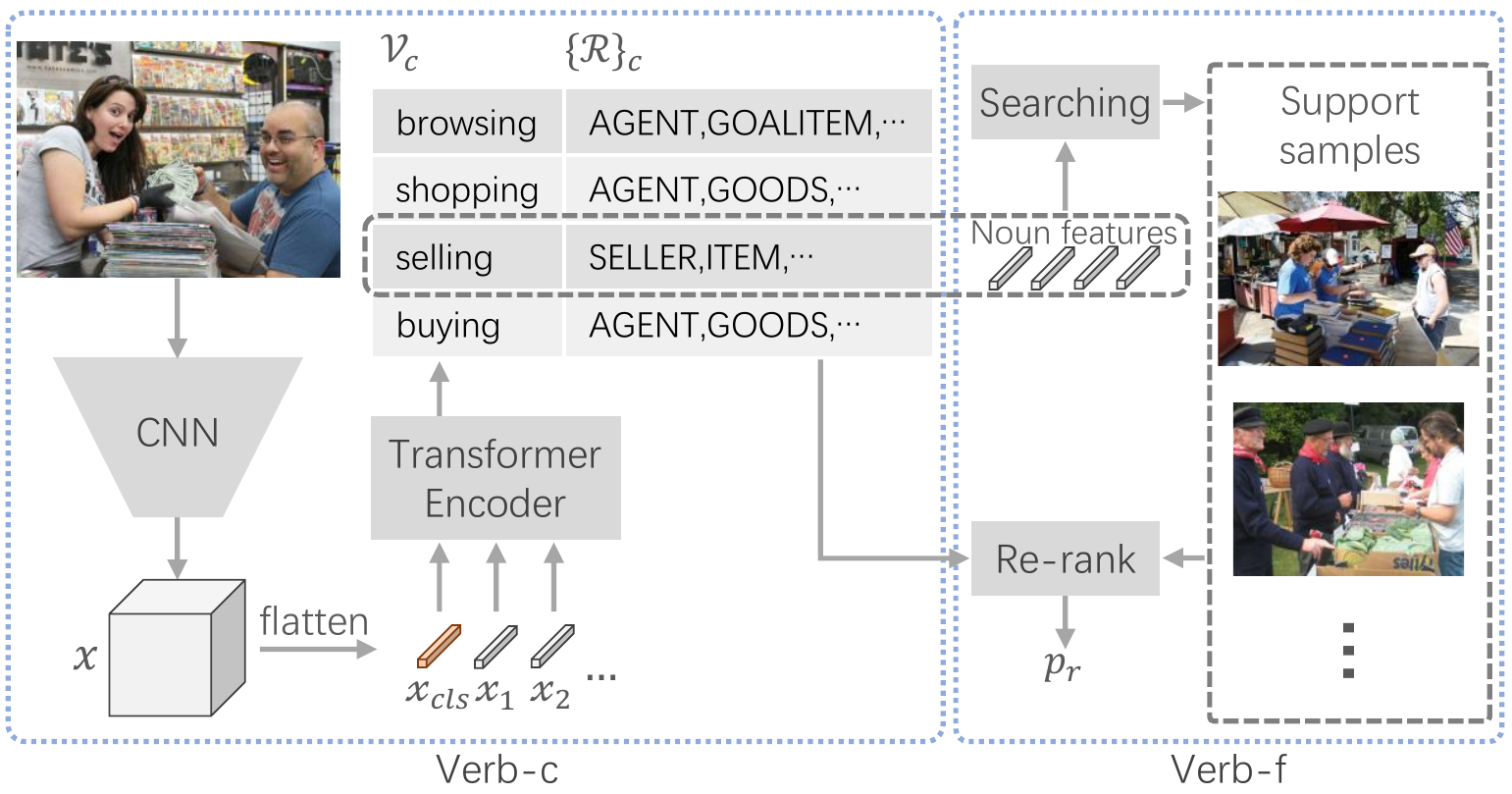}
            \caption{The architecture of CFVM, including Verb-c (left) and Verb-f (right).}
            \label{fig:verb_model}
        \end{minipage}
    \end{minipage}
\end{figure*}

\subsection{Transformer-based Noun Model (TNM)}

The noun model TNM is designed to detect (\ie, classify and ground) all semantic roles of a verb frame. Inspired from recently proposed end-to-end transformer-based object detector DETR~\cite{carion2020end}, TNM is also a transformer-based model. As shown in Figure~\ref{fig:noun_model}, TNM consists of four sub-networks: a CNN backbone, a transformer encoder, a transformer decoder and a noun detection head.

Given image $I$, the CNN backbone first extracts a feature map $\bm{X}^N \in\mathbb{R}^{C \times {H} \times {W}}$. Since the input for the transformer encoder is a sequence of tokens, the feature map $\bm{X}^N$ is flattened to a sequence of ``visual" tokens: $[\bm{x}^N_1, ..., \bm{x}^N_{H*W}]$, and each token $\bm{x}^N_i \in \mathbb{R}^C$ is a C-dim visual feature. Then, the visual token sequence is fed into the transformer encoder:
\begin{equation}
    \left[ \tilde{\bm{x}}^N_1, ..., \tilde{\bm{x}}^N_{H*W} \right] = F^{\text{TNM}}_{\text{enc}} (\left[\bm{x}^N_1, ..., \bm{x}^N_{H*W} \right]),
\end{equation}
where $F^{\text{TNM}}_{\text{enc}}$ is a vanilla transformer encoder, which consist of a position embedding layer, and a set of stacked multi-head self-attention layers. We refer the readers to the original Transformer~\cite{vaswani2017attention} paper for more details.


Given the encoded visual feature $\tilde{\bm{X}}^N = [\tilde{\bm{x}}^N_1, ..., \tilde{\bm{x}}^N_{H*W} ]$, verb $v$ and its semantic role set $\mathcal{R}=\{r_1,...,r_m\}$, we first encode the verb $v$ and all semantic roles $\{r_i\}$ into query embeddings: $\bm{q}_v$ and $\{\bm{q}_{r_i}\}$. Then, these query embeddings are regarded as the input queries for the transformer decoder (\cf~Figure~\ref{fig:noun_model}), and the encoded visual feature $\tilde{\bm{X}}^N$ is the key and value for the cross-attention layer in the decoder, \ie, 
\begin{equation} \label{equ:4}
    [\bm{x}_v, \bm{x}_{n_1}, ..., \bm{x}_{n_m}] = F^{\text{TNM}}_{\text{dec}} (\tilde{\bm{X}}^N, [\bm{q}_v, \bm{q}_{r_1}, ..., \bm{q}_{r_m}]),
\end{equation}
where $\bm{x}_v$ and $\bm{x}_{n_i}$ are the output of query $\bm{q}_v$ and $\bm{q}_{r_i}$, respectively. Lastly, a lightweight noun detection head (MLP) predicts the object category and regresses the normalized bounding box coordinates for each semantic role query, \ie, 
\begin{equation}
    \{n_i, b_i\} = \text{MLP}(\bm{x}_{n_i}),
\end{equation}
and the output of TNM is the set of object categories and bounding boxes of all roles, \ie, $\mathcal{N}$ and $\mathcal{B}$ (\cf~Eq.~\eqref{equ:2}).

\noindent\textbf{Differences with DETR.} Although TNM has a similar architecture with DETR, there are several notable differences: 1) \emph{Meaning of decoder input queries.} For DETR, these queries can be regarded as priors for potential objects with different sizes and locations. Thus, the query number should be large (\eg, 100 queries for COCO). Instead, each query in TNM is the embedding of a semantic role, which is responsible for detecting this specific role, and the maximum number of semantic roles is small (\eg, 6 queries for SWiG). 2) \emph{Matching algorithm for optimization.} For DETR, a matching algorithm is needed for optimal bipartite matching between ground-truth and set predictions. While in TNM, there is a perfect one-to-one match for each role query. Thanks to this design, TNM can not only take advantage of the prior knowledge of the verb but also reduce computational complexity.



\subsection{Coarse-to-Fine Verb Model (CFVM)}

CFVM is a two-step verb classification model, which consists of two modules: 1) a coarse-grained verb model (Verb-c) to propose a set of verb candidates; 2) a fine-grained verb model (Verb-f) to make the final verb prediction. The overview architecture of CFVM is illustrated in Figure~\ref{fig:verb_model}.

\noindent\textbf{Coarse-grained Verb Classification.}
The coarse-grained verb classification model (Verb-c) aims to propose a set of verbs as initial candidates. Since the verb (or activity) classification task inherently needs to model the semantic relationships between multiple objects in the image, we combine a CNN backbone with a transformer encoder as our Verb-c (\cf~Figure~\ref{fig:verb_model} (left)), \ie, the self-attention mechanism in the transformer encoder helps to capture global context in the image. Instead, almost all existing GSR (or SR) methods directly use a plain CNN backbone (\eg, ResNet or VGG) as their verb classification model.


Similarly to the TNM, for given image $I$, the CNN backbone first extracts feature map $\bm{X}^V \in \mathbb{R}^{C\times H\times W}$, and $\bm{X}^V$ is flatten to a sequence of tokens: $[\bm{x}^V_1, ..., \bm{x}^V_{H\times W}]$. Following the convention of BERT-family works~\cite{devlin2018bert}, we also add a learnable embedding $\bm{x}_{cls}$ of special token [\texttt{CLS}] to the token sequence. Then, this augmented token sequence is fed into the transformer encoder $F^{\text{Verb-c}}_{\text{enc}}$:
\begin{equation} \label{equ:6}
    \left[ \tilde{\bm{x}}_{cls}, \tilde{\bm{x}}^V_1, \tilde{\bm{x}}^V_2, ... \right] = F^{\text{Verb-c}}_{\text{enc}} (\left[\bm{x}_{cls}, \bm{x}^V_1, \bm{x}^V_2, ... \right]).
\end{equation}
The encoded embedding of [\texttt{CLS}] token (\ie, $\tilde{\bm{x}}_{cls}$) is used to represent the gist of the whole image, and it is fed into a fully-connected layer to make the coarse verb prediction. We select top-$N$ verbs as candidates and denote them as $\mathcal{V}_c$.

\noindent\textbf{Fine-grained Verb Classification.} Up to now, for image $I$, we have obtained the top-$N$ candidate verbs $\mathcal{V}_c$ and semantic role set $\{\mathcal{R}\}_c$ from Verb-c. Then, for each candidate $v_i \in \mathcal{V}_c$ and $\mathcal{R}_i \in \{\mathcal{R}\}_c$, we can also get the corresponding semantic role detection results from noun model TNM:
\begin{equation}
    \mathcal{N}_i, \mathcal{B}_i = \text{TNM} (v_i, \mathcal{R}_i, I).
\end{equation}
Thus, we obtain the semantic role detection results for all $N$ verb candidates: $\{\mathcal{N}\}_c$ and $\{\mathcal{B}\}_c$.

To determine the final verb prediction, we first retrieve $M$ support images from the training set for each verb candidate $v_i \in \mathcal{V}_c$. The support image set for $v_i$ is denoted as $\mathcal{I}_i = \{I^{(i)}_1, ..., I^{(i)}_M \}$. Each support image $I^{(i)}_j$ is retrieved based on the semantic role feature similarity scores $S(I, I_k)$, \ie,
\begin{equation}
\mathcal{I}_i = \arg\text{top-M}_{I_k \in \mathcal{D}_i} S(I, I_k),
\end{equation}
where $\mathcal{D}_i$ is the set of all training set images with ground-truth verb category $v_i$. The semantic role feature similarity score $S(\cdot)$ is the average cosine similarity of all roles:
\begin{equation}
    S(I, I_k) = \frac{1}{m} \sum_{i=1}^{m} sim(\bm{x}_{n_i}^{I},\bm{x}_{n_i}^{I_k}),
\end{equation}
where $\bm{x}_{n_i}^{I}$ and $\bm{x}_{n_i}^{I_k}$ is the $i$-th semantic role features from the output of TNM (\cf~Eq.~\eqref{equ:4}) of image $I$ and $I_k$, respectively. Similarity function $sim$ is cosine similarity.



After retrieving support image set $\mathcal{I}_i$ for each verb candidate $v_i$, our fine-grained verb model (Verb-f) uses a lightweight MLP $\phi(\cdot)$ to map the original coarse verb feature $\tilde{\bm{x}}_{cls}$ (\cf~Eq.~\eqref{equ:6}) into a more distinctive embedding space, \ie, $\phi$ is designed to project coarse verb features of image $I$ and all retrieved support images to focus on fine distinctive details. (We train $\phi(\cdot)$ with hard triplets, and the training details are in the following sections).

In the inference stage, given the coarse-grained classification scores $\{p(v_1),p(v_2),...,p(v_N)\}$ of all top-$N$ candidates, the Verb-f model re-ranks all verb candidates based on the similarity scores between image $I$ and support images. If $p(v_1) >= \epsilon$ ($\epsilon$ is a threshold score), the final verb prediction is $v_1$, \ie, the original Verb-c model performs well. If $p(v_1) < \epsilon$, the re-ranked scores $p^r(v_i)$ is calculated as:
\begin{equation}
\begin{aligned}
p^r(v_i) = \beta \sum_{I_k \in \mathcal{I}_i} sim(\phi(\tilde{\bm{x}}_{cls}^I), \phi(\tilde{\bm{x}}_{cls}^{I_k})) * S(I, I_k) + \alpha p(v_i),
\end{aligned}
\end{equation}
where $\alpha$ and $\beta$ are weights for the trade-off between original verb prediction probability $p(v_i)$ from Verb-c and the confidence from support image set in Verb-f.













\subsection{Training Objectives}
\label{sec:training}

In the training stage, we train all components in SituFormer separately, including TNM, Verb-c, and Verb-f:

\noindent\textbf{Training Objective of TNM.}
We denote the ground-truth noun categories and bounding boxes as $\mathcal{N}^{gt}, \mathcal{B}^{gt}$ and the predicted noun categories and bounding boxes as $\hat{\mathcal{N}},\hat{\mathcal{B}}$. The detection loss $\mathcal{L}_{\text{TNM}}$ of TNM is calculated as:
\begin{equation}
   \mathcal{L}_{\text{TNM}} = \sum_{i=1}^{m} \left[\text{XE}(n^{gt}_i, \hat{n}_i) + \mathcal{L}_{box}(b^{gt}_i, \hat{b}_i)\right],
\end{equation}
where XE is the cross-entropy loss and $\mathcal{L}_{box}$ consists of a generalize IoU loss~\cite{rezatofighi2019generalized} and a L1 regression loss.

\noindent\textbf{Training Objective of Verb-c.}
We denote the ground-truth verb category as $v^{gt}$ and the predicted verb category as $\hat{v}$.The classification loss of Verb-c $\mathcal{L}_{\text{verb-c}}$ is calculated as:
\begin{equation}
    \mathcal{L}_{\text{verb-c}} = \text{XE}(v^{gt}, \hat{v}).
\end{equation}

\noindent\textbf{Training Objective of Verb-f.}
For each training sample $I^a$ (anchor image), we regard all support images with the same ground-truth verb category as hard positive sample set $\mathcal{I}^+$, and all support images for other verb categories as hard negative sample set $\mathcal{I}^-$, \ie, $\mathcal{I}^- = \{\mathcal{I}_i\}\setminus \mathcal{I}^+$. The margin-based triplet loss of Verb-f $\mathcal{L}_{\text{verb-f}}$ is calculated as: 
\begin{equation}
\begin{aligned}
    \mathcal{L}_{\text{verb-f}} = \max (0, \tau & +  sim(\phi(\tilde{\bm{x}}_{cls}^{I^{a}}), 
    \phi (\tilde{\bm{x}}_{cls}^{X_{i}^{n}}))  \\
    & - sim(\phi(\tilde{\bm{x}}_{cls}^{I^{a}}), \phi (\tilde{\bm{x}}_{cls}^{X_{i}^{p}}))),
\end{aligned}
\end{equation}
where $X_{i}^{p} \in \mathcal{I}^{+} $ is the hard positive image and  $X_{i}^{n} \in \mathcal{I}^{-} $ is the hard negative image. $\tau$ is a margin value.

\addtolength{\tabcolsep}{-3pt}
\addtolength{\abovecaptionskip}{-10pt}
\begin{table*}[tb]
    \begin{center}
    \scalebox{1.0}{
    \begin{tabular}{l|ccccc|ccccc|cccc}
    \hline
    \multirow{2}{*}{Models} & \multicolumn{5}{c|}{Top-1-Verb} & \multicolumn{5}{c|}{Top-5-Verb} & \multicolumn{4}{c}{Ground-Truth-Verb} \\
    & verb & value & val-all & grnd & grnd-all & verb & value & val-all & grnd & grnd-all & value & val-all & grnd & grnd-all \\
    \hline
    \multicolumn{15}{l}{\emph{Situation Recognition Models}} \\
    \hline
    CRF & 32.25 & 24.56 & 14.28 & -- & -- & 58.64 & 42.68 & 22.75 & -- & -- & 65.90 & 29.50 & -- & -- \\
    CRF+DataAug & 34.20 & 25.39 & 15.61 & -- & -- & 62.21 & 46.72 & 25.66 & -- & -- & 70.80 & 34.82 & -- & -- \\
    VGG+RNN & 36.11 & 27.74 & 16.60 & -- & -- & 63.11 & 47.09 & 26.48 & -- & -- & 70.48 & 35.56 & -- & -- \\
    FC-Graph & 36.93 & 27.52 & 19.15 & -- & -- & 61.80 & 45.23 & 29.98 & -- & -- & 68.89 & 41.07 & -- & -- \\
    CAQ & 37.96 & 30.15 & 18.58 & -- & -- & 64.99 & 50.30 & 29.17 & -- & -- & 73.62 & 38.71 & -- & -- \\
    Kernel-Graph & 43.21 & 35.18 & 19.46 & -- & -- & 68.55 & 56.32 & 30.56 & -- & -- & 73.14 & 41.68 & -- & -- \\
    \hline
    \multicolumn{15}{l}{\emph{Grounded Situation Recognition Models}} \\
    \hline
    ISL & 38.83 & 30.47 & 18.23 & 22.47 & 7.64 & 65.74 & 50.29 & 28.59 & 36.90 & 11.66 & 72.77 & 37.49 & 52.92 & 15.00 \\
    JSL & 39.60 & 31.18 & 18.85 & 25.03 & 10.16 & 67.71 & 52.06 & 29.73 & 41.25 & 15.07 & 73.53 & 38.32 & 57.50 & 19.29 \\
    \cline{1-1}
    \multirow{2}{*}{\makecell[l]{\textbf{SituFormer} \\ \small{Gains ($\Delta$)} }} & \textbf{44.32} & \textbf{35.35} & \textbf{22.10} & \textbf{29.17} & \textbf{13.33} & \textbf{71.01} & \textbf{55.85} &
    \textbf{33.38} & \textbf{45.78} & \textbf{19.77} & \textbf{76.08} & \textbf{42.15} & \textbf{61.82} & \textbf{24.65} \\
    & \small{+4.72} & \small{+4.17} & \small{+3.25} & \small{+4.14} & \small{+3.17} & \small{+3.30} & \small{+3.79} & \small{+3.65} & \small{+4.53} & \small{+4.70} & \small{+2.55} & \small{+3.83} & \small{+4.32} & \small{+5.36}  \\
    \hline
    \end{tabular}
    }
    \end{center}
    \caption{Performance (\%) of state-of-the-art GSR (and SR) methods on SWiG dataset development (dev) set.}
    \label{tab:tab_performance_dev}
\end{table*}
\addtolength{\abovecaptionskip}{10pt}
\addtolength{\tabcolsep}{3pt}

\addtolength{\tabcolsep}{-3pt}
\addtolength{\abovecaptionskip}{-10pt}
\begin{table*}[tb]
    \begin{center}
    \scalebox{1.0}{
    \begin{tabular}{l|ccccc|ccccc|cccc}
    \hline
    \multirow{2}{*}{Models} & \multicolumn{5}{c|}{Top-1-Verb} & \multicolumn{5}{c|}{Top-5-Verb} & \multicolumn{4}{c}{Ground-Truth-Verb} \\
    & verb & value & val-all & grnd & grnd-all & verb & value & val-all & grnd & grnd-all & value & val-all & grnd & grnd-all \\
    \hline
    ISL & 39.36 & 30.09 & 18.62 & 22.73 & 7.72 & 65.51 & 50.16 & 28.47 & 36.60 & 11.56 & 72.42 & 37.10 & 52.19 & 14.58 \\
    JSL & 39.94 & 31.44 & 18.87 & 24.86 & 9.66 & 67.60 & 51.88 & 29.39 & 40.60 & 14.72 & 73.21 & 37.82 & 56.57 & 18.45 \\
    \cline{1-1}
    \multirow{2}{*}{\makecell[l]{\textbf{SituFormer} \\ \small{Gains ($\Delta$)} }} & \textbf{44.20} & \textbf{35.24} & \textbf{21.86} & \textbf{29.22} & \textbf{13.41} & \textbf{71.21} & \textbf{55.75} & \textbf{33.27} & \textbf{46.00} & \textbf{20.10} & \textbf{75.85} & \textbf{42.13} & \textbf{61.89} & \textbf{24.89} \\
    & \small{+4.26} & \small{+3.80} & \small{+2.99} & \small{+4.36} & \small{+3.75} & \small{+3.61} & \small{+3.87} & \small{+3.88} & \small{+5.40} & \small{+5.38} & \small{+2.64} & \small{+4.31} & \small{+5.32} & \small{+6.44}  \\
    \hline
    \end{tabular}
    }
    \end{center}
    \caption{Performance (\%) of state-of-the-art GSR methods on SWiG dataset test set.}
    \label{tab:tab_performance_test}
\end{table*}
\addtolength{\abovecaptionskip}{10pt}
\addtolength{\tabcolsep}{3pt}

\section{Experiments}
\subsection{Experimental Settings}

\noindent \textbf{Datasets.}
We evaluated our method for GSR on the challenging \textit{SWiG} benchmark~\cite{pratt2020grounded}. It is an extension dataset of the SR dataset \textit{imSitu}~\cite{yatskar2016situation}. Specifically, each image in \textit{imSitu} is annotated with three verb frames by three annotators. \textit{SWiG} adds bbox annotations for all visible semantic roles ($63.9\%$ roles have bbox annotations). \textit{SWiG} consists of $126,102$ images with $9,928$ object categories, $190$ semantic role types and $504$ verb categories. The official splits are $75$K/$25$K/$25$K images for training, validation, and test set, respectively. 

\noindent \textbf{Evaluation Metrics.}
We followed prior work~\cite{pratt2020grounded} to evaluate our method on five metrics: 1) \textbf{verb}: The accuracy of verb prediction. 2) \textbf{value}: The accuracy of noun prediction w.r.t each semantic role. 3) \textbf{value-all (val-all)}: The accuracy of noun prediction w.r.t. the whole semantic role set. 4) \textbf{grounded-value (grnd)}: The accuracy of noun prediction with correct grounding w.r.t each semantic role. By ``correct grounding", we mean the IoU between predicted bounding box and ground-truth is large than threshold $0.5$. 5) \textbf{grounded-value-all (grnd-all)}: The accuracy of noun prediction with correct grounding w.r.t to the whole semantic role set. Meanwhile, there are three different evaluation settings: 1) \textbf{\texttt{Ground-Truth-Verb}}: The ground-truth verbs is assumed to be known. 2) \textbf{\texttt{Top-1-Verb}}: \emph{verb} reports the top-$1$ accuracy, and all other four \emph{value} metrics are considered wrong if verb is wrong. 3) \textbf{\texttt{Top-5-Verb}}: \emph{verb} reports the top-$5$ accuracy, and all other four \emph{value} metrics are conditioned on the correct verb having been predicted.


\noindent \textbf{Implementation Details.}
The CNN backbone of both TNM and CFVM were ResNet-50 pretrained on ImageNet. The decoder of TNM used sine position encodings. For TNM, we followed DETR and set the layer number of the encoder and decoder as $6$ by default except as otherwise noted. Following prior works, TNM only predicted the top $2,000$ most frequent object categories, which covers about $95\%$ noun annotations. TNM was trained with AdamW optimizer and the initial learning rate of transformer and CNN backbone was set to $10^{-4}$ and $10^{-5}$ respectively. We trained it for $20$ epoch with a learning rate drop by a factor of $10$ after $10$ epoch on $4$ V100 GPUs. The total batch size was set to $128$. Verb-c model had the same training strategy with TNM. For Verb-f model, the hard negative sample set was constructed from the top-$5$ verb candidates, and the size of support image set of each verb was set to $10$.
At each training step, we randomly chose a negative sample and a positive sample to compose the training triplet. Verb-f was trained for $20$ epoch with an initial learning rate $5 \times 10^{-4}$ drop by a factor of $10$ after $10$ epoch. The margin $\tau$ was set to $0.2$. In the inference stage, the threshold score $\epsilon=0.4$ and weights $\alpha = \beta = 0.5$.






\subsection{Comparisons with State-of-the-Arts}
\noindent\textbf{Settings.} We compared our SituFormer with state-of-the-art GSR and SR models on SWiG dataset. Based on their model architectures, existing SR models are be coarsely grouped into: 1) CRF-based models: \textbf{CRF}~\cite{yatskar2016situation} and \textbf{CRF+DataAug}~\cite{yatskar2017commonly}. 2) RNN-based models: \textbf{VGG+RNN}~\cite{mallya2017recurrent}. 3) GNN-based models: \textbf{FC-Graph}~\cite{li2017situation}, \textbf{Kernel-Graph}~\cite{suhail2019mixture}. 4) Attention-based: \textbf{CAQ}~\cite{9156513}. For GSR, all existing model: \textbf{ISL} and \textbf{JSL}~\cite{pratt2020grounded} are RNN-based. The results on the \textit{development} (dev) set and \emph{test} set are illustrated in Table~\ref{tab:tab_performance_dev} and Table~\ref{tab:tab_performance_test}, respectively.

\noindent\textbf{Results under \texttt{Ground-Truth-Verb} Setting.} Under this setting, we can evaluate the model performance on semantic role detection (\ie, TNM). Based on results on Table~\ref{tab:tab_performance_dev} and Table~\ref{tab:tab_performance_test}, we can have the following observations: 1) For role classification (\ie, \emph{value} and \emph{val-all} metrics), SituFormer outperforms all existing GSR (and SR) models on both metrics. Compared to the best performer Kernel-Graph, we achieve $2.94\%$ ($76.08\%$ vs. $73.14\%$) and $0.47\%$ absolute performance gains under \emph{value} and \emph{val-all} metrics (on the dev set), respectively. 2) As for the grounding metrics (\ie, \emph{grnd} and \emph{grnd-all} metrics), SituFormer also outperforms all existing GSR models. Compared to JSL, performance gains are much more significant, \eg, $5.36\%$ ($24.65\%$ vs. $19.29\%$) and $6.44\%$ ($24.89\%$ vs. $18.45\%$) absolute performance gains under \emph{grnd-all} metric on dev and test set, respectively. 


\noindent\textbf{Results under \texttt{Top-N-Verb} settings.} From the \emph{verb} metric, we can observe that SituFormer outperforms all existing GSR (and SR) models on both top-$1$ and top-$5$ verb accuracy, which demonstrate the superiority of CFVM. With the SOTA results of both TNM and CFVM, SituFormer also achieves the best results on \emph{val-all}, \emph{grnd} and \emph{grnd-all} under this setting. Although SR model Kernel-Graph outperforms SituFormer slightly on the \textit{value} metric (\ie, $0.47\%$ under \texttt{Top-5-Verb} setting), they actually significantly sacrifice their performance on \textit{val-all} metric due to the joint training of their verb model and noun model.


\subsection{Ablation Studies}
We conducted extensive ablation studies to demonstrate the effectiveness of each component of our Situformer.

\noindent\textbf{Effectiveness and Hyper-parameters Choices of CFVM.} 

\noindent\emph{Effectiveness of Coarse-to-Fine Classification.}
To validate the effectiveness of Verb-f, we conducted ablations by using only Verb-c as the verb model and TNM as the noun model (\ie, denoted as ``SituFormer w/o Verb-f"). The results under \texttt{Top-1-Verb} setting are reported in Table~\ref{tab:ablation_rerank}. From the results, we can observe that the Verb-f model (\ie, the coarse-to-fine strategy) can directly improve the final top-1 verb accuracy by $1.12\%$. Accordingly, all value-related (\ie, \emph{value}, \emph{val-all}, \emph{grnd}, \emph{grnd-all}) metrics are further boosted. 

\addtolength{\tabcolsep}{-3pt}
\addtolength{\abovecaptionskip}{-5pt}
\begin{table}[t]
    \centering
    \scalebox{0.95}{
    \begin{tabular}{l|ccccc}
        \hline
        Models & verb & value & val-all & grnd & grnd-all \\
        \hline
        SituFormer w/o Verb-f & 43.08 & 34.20 & 21.24 & 28.45 & 12.90 \\
        \textbf{SituFormer} & 44.20 & 35.24 & 21.86 & 29.22 & 13.41  \\
        \hline
    \end{tabular}
    }
    \caption{Performance (\%)  under \texttt{Top-1-Verb} setting.}
    \label{tab:ablation_rerank}
\end{table}
\addtolength{\abovecaptionskip}{5pt}
\addtolength{\tabcolsep}{3pt}

\addtolength{\abovecaptionskip}{-10pt}
\begin{figure*}
    \centering
    \includegraphics[width=1.0\linewidth]{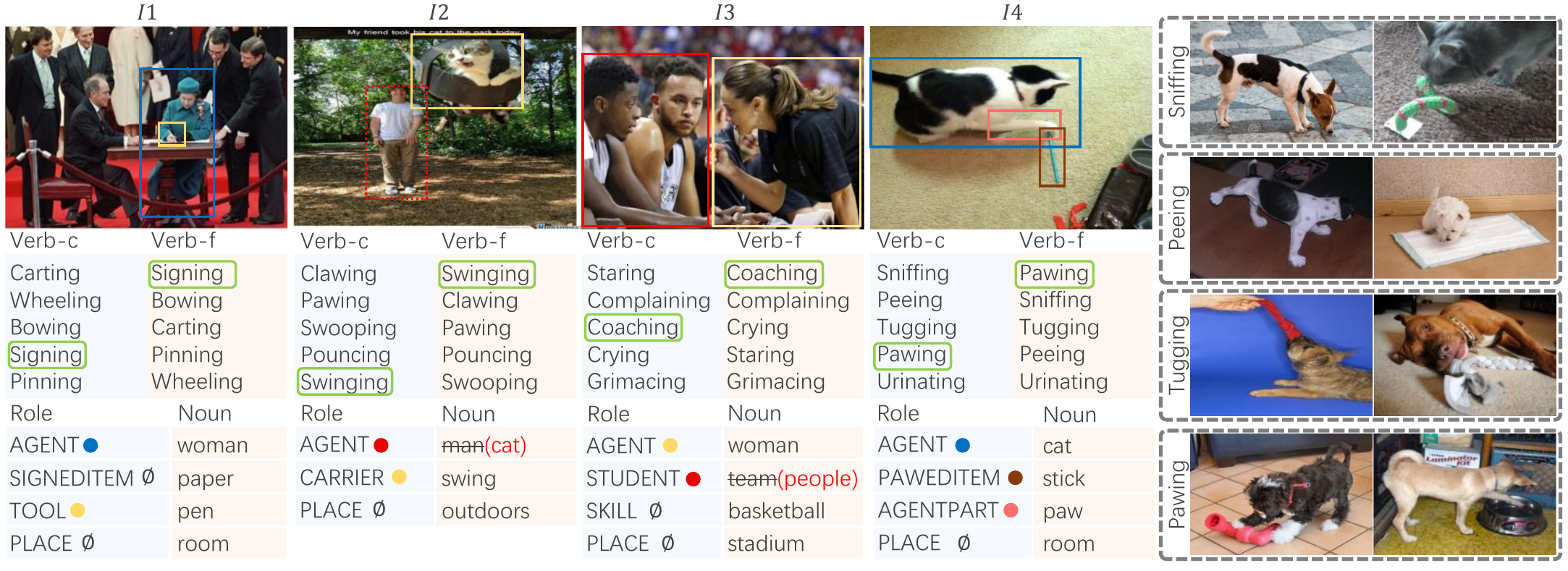}
    \caption{\textbf{Left}: For each image, the top-$5$ verb candidates and re-ranked verbs are shown below ``Verb-c" and ``Verb-f", respectively. The semantic role detection are shown in the third row. Incorrectly object category prediction is scratched out with ground-truth shown in red brackets. The correct groundings are shown in solid boxes while the incorrect ones shown in dotted boxes. $\emptyset$ means no ground-truth grounding for that role. \textbf{Right}: The retrieved support images of top-$4$ verb candidates for $I$4.}
    \label{fig:viz}
\end{figure*}
\addtolength{\abovecaptionskip}{10pt}

\begin{figure*}[!h]
    \begin{minipage}[c]{0.60\linewidth}
        \addtolength{\abovecaptionskip}{-8pt}
        \captionsetup{type=table} 
        	\subfloat[Top-1 \& top-5 verb accuracy w.r.t. different layer numbers of the encoder in Verb-c (\ie, w/o Verb-f).]{
        		\tablestyle{2.5pt}{1.05}\begin{tabular}{c|x{22}x{22}|x{35}|x{22}x{22}}
        			\hline
        			\# Layers & Top-1 & Top-5 & \# Layers & Top-1 & Top-5  \\
        		    \hline
        			0(JSL) & 39.94 & 67.60 & 4 & \textbf{43.08} & \textbf{71.21} \\
                    1 & 41.88 & 70.10 & 5 & 42.74 & 71.22 \\
                    2 & 42.52 & 70.95 & 6 & 42.34 & 70.56 \\
                    3 & 42.79 & 70.93 & & & \\
        			\hline
        	\end{tabular}}\hspace{3mm}
        	\subfloat[Verb accuracy w.r.t. sizes of support images.]{
        		\tablestyle{2.5pt}{1.05}\begin{tabular}{c|x{22}}
        			\hline
        			\# Support
        			Imgs & Top-1  \\
        			\hline
                    1 & 43.50 \\
                    3 & 43.96 \\
                    5 & \textbf{44.19} \\
                    10 & \textbf{44.20} \\
        			\hline
        	\end{tabular}}
        \caption{Results (\%) on different hyper-parameters choices of CFVM.}
        \label{table:ablative_CFVM}
        \addtolength{\abovecaptionskip}{8pt}
    \end{minipage} \hfill
    \begin{minipage}[c]{0.4\linewidth}
        \captionsetup{type=table} 
		    \tablestyle{0.5pt}{1.05}\begin{tabular}{c|c|x{25}x{25}x{25}x{30}}
			\hline
            V-query & Shared R-query & value & val-all & grnd & grnd-all \\
            \hline
            \Checkmark & \Checkmark  & \textbf{75.85} & \textbf{42.13} & \textbf{61.89} & \textbf{24.89}  \\
            \XSolidBrush & \Checkmark & 74.17 & 39.37 & 60.16 & 22.86 \\
            \Checkmark & \XSolidBrush & 73.26 & 38.13 & 57.02 & 20.42 \\
            \XSolidBrush & \XSolidBrush & 70.96 & 34.87 & 55.37 & 19.02 \\			\hline
        	\end{tabular}
          \caption{Results (\%) of different query designs in TNM under \texttt{Ground-Truth-Verb} setting.}
          \label{tab:ablative_TNM}
    \end{minipage}
\end{figure*}

\noindent\emph{Layer Numbers of the Encoder in Verb-c.} We investigated verb accuracy (both top-1 and top-5) of Verb-c with different layer numbers of transformer encoder (up to 6), and the results are reported in Table~\ref{table:ablative_CFVM} (a). The baseline model (denoted as 0 layer) is the ResNet-50 model, which is the same verb model used in JSL. From Table~\ref{table:ablative_CFVM} (a), we can observe that applying the transformer encoder can gradually improve the verb accuracy (\eg, 43.08\% vs. 39.94\%). And when the stacked layer number more than 4 layers, their performances reach the plateaus. To trade-off between accuracy and computation, we used four encoder layers in our Verb-c model.



\noindent\emph{The Size of Support Images Set in Verb-f.}
We explored various support image set sizes to show the robustness of the retrieve-and-rerank scheme of the verb-f, and we reported the top-1 verb accuracy in Table~\ref{table:ablative_CFVM} (b). From the results, we can observe that larger support image set size can perform better but the accuracy plateaus when $n > 5$. 
This is because the possible number of hard support image is limited. 

\noindent\textbf{Query Designs in TNM.} Since one of key differences between TNM and DETR-family models is the design of decoder queries, we also investigated several different designs:

\noindent\emph{Importance of Verb Query (V-queries).} Since the verb itself provide useful inductive bias for semantic role prediction, it is intuitive that introducing auxiliary verb query is helpful. To validate the effectiveness of the verb query, we conducted ablations under the \texttt{Ground-Truth-Verb} setting, and the results are reported in Table~\ref{tab:ablative_TNM}. From the results, we can observe that the verb query brings $1.68\%$ and $2.30\%$ absolute performance gains on the \emph{value} metric. It is also worth noting that TNM without verb query already achieves new state-of-the-art performance on all four metrics.


\noindent\emph{Effectiveness of Sharing Role Queries (R-queries).} To mitigate semantic sparsity issue (\ie, numerous triplets $\langle$\texttt{verb} - \texttt{role} - \texttt{noun} $\rangle$ are rare in the SWiG dataset), TNM shares the role embeddings as queries among all different verbs. To validate the effectiveness of sharing role queries, we conducted ablations by using TNM without sharing role queries (\ie, each verb has independent semantic role queries). Results under \texttt{Ground-Truth-Verb} setting are reported in Table~\ref{tab:ablative_TNM}. From the results, we can observe that the improvement of sharing role queries is obvious (\eg, $\approx 2\%$ and  $3\%$ performance gains for the \emph{value} and \emph{val-all} metrics.

\noindent\emph{Effectiveness of parallel decoding.} As shown in Table~\ref{tab:tab_performance_dev} \& Table~\ref{tab:tab_performance_test} Ground-Truth-Verb setting where the effect of verb model is exempted, the quantitative improvements to ISL and JSL attest to the effectiveness of parallel decoding.

\noindent\textbf{Qualitative Results.}
In Figure~\ref{fig:viz}, we display coarse-to-fine verb predictions and semantic role detection results of some images (\textit{I}1 
$\sim$ \textit{I}4) from the test set (left) and retrieved support image sets (right). For all four examples, their initial top-$1$ verb predictions from Verb-c are wrong but the ground-truth verbs are probabilities ascend to the 1st place after re-ranking by Verb-f. 
We can see that discriminative details are needed to distinguish ground-truth verb from these candidates (\eg, tiny interaction between the cat and the stick in \textit{I}4).
We also display some errors of TNM.
In \textit{I}2, the \texttt{AGENT} of \texttt{Swinging} is incorrectly predicted as \texttt{man}. This error may be caused by the rare occurrence of ``a cat is swinging".
While in \textit{I}3, the incorrect \texttt{team} for \texttt{STUDENT} is actually more reasonable than the ground-truth \texttt{people}. On the right part of Figure~\ref{fig:viz}, we show retrieved support images of \texttt{Pawing} and the top-3 wrong predicted verbs.




\section{Conclusions}
In this paper, we argue that the existing two-stage GSR models have drawbacks in both verb prediction stage (insufficient to handle high diversity of daily activities) and semantic role detection stage (using autoregressive model). To alleviate these drawbacks, we propose 
SituFormer which consists of a two-step verb model and a transformer-based noun model. Specifically, the verb model predicts verb in a coarse-to-fine process, and the noun model makes use of the flexibility of transformer to integrate the recognition and grounding of roles. We achieved significant gains over all metrics on challenging benchmark \textit{SWiG}, and conducted ablative analysis for each component of SituFormer.

\section{Acknowledgements}
This research is funded by Sea-NExT Joint Lab, Singapore.

\bibliography{aaai22}

\end{document}